\documentclass[letterpaper, 10 pt, journal, twoside]{IEEEtran}
\usepackage{graphics} 
\usepackage{epsfig} 
\usepackage{times} 
\usepackage{amsmath} 
\usepackage{amssymb}  
\usepackage{multirow}
\usepackage{booktabs}
\usepackage{float} 
\usepackage{xcolor}
\usepackage{gensymb}
\usepackage{algorithm}
\usepackage[noend]{algpseudocode}
\usepackage[english]{babel}

\usepackage{breqn}

\usepackage{cite}

\ifCLASSINFOpdf
\else
\fi
\begin{document}
%
\title{Lifelong 3D Mapping Framework for Hand-held \& Robot-mounted LiDAR Mapping Systems}
%
%
%

\author{Liudi Yang$^{*}$, Sai Manoj Prakhya$^{*}$, Senhua Zhu and Ziyuan Liu

\thanks{Manuscript received: March, 19, 2024; Revised April, 10, Year; Accepted June, 12, 2024.}
\thanks{This paper was recommended for publication by Editor Javier Civera upon evaluation of the Associate Editor and Reviewers' comments.}
\thanks{$^{*}$denotes equal contribution.}
\thanks{Liudi Yang, Sai Manoj Prakhya \& Ziyuan Liu are part of the Intelligent Cloud Technologies Lab, Huawei, Munich, Germany. Senhua Zhu is the Director of Algorithm Innovation lab at Huawei Cloud, China.}%
\thanks{Digital Object Identifier (DOI): see top of this page.}
}
%
%

\markboth{IEEE Robotics and Automation Letters. Preprint Version. Accepted June, 2024}
{yang \MakeLowercase{\textit{et al.}}: Lifelong 3D Mapping Framework for Hand-held \& Robot-mounted LiDAR Mapping Systems} 

%



\maketitle

\begin{abstract}
We propose a lifelong 3D mapping framework that is modular, cloud-native by design and more importantly, works for both hand-held and robot-mounted 3D \mbox{LiDAR} mapping systems. Our proposed framework comprises of dynamic point removal, multi-session map alignment, map change detection and map version control. First, our sensor-setup agnostic dynamic point removal algorithm works seamlessly with both hand-held and robot-mounted setups to produce clean static 3D maps. Second, the multi-session map alignment aligns these clean static maps automatically, without manual parameter fine-tuning, into a single reference frame, using a two stage approach based on feature descriptor matching and fine registration. Third, our novel map change detection identifies positive and negative changes between two aligned maps. Finally, the map version control maintains a single base map that represents the current state of the environment, and stores the detected positive and negative changes, and boundary information. Our unique map version control system can reconstruct any of the previous clean session maps and allows users to query changes between any two random mapping sessions, all without storing any input raw session maps, making it very unique. Extensive experiments are performed using hand-held commercial LiDAR mapping devices and open-source robot-mounted LiDAR SLAM algorithms to evaluate each module and the whole 3D lifelong mapping framework.\end{abstract}

\begin{IEEEkeywords}
Lifelong Mapping, Dynamic Object Removal, Map Alignment, Map Version Control
\end{IEEEkeywords}

%
\IEEEpeerreviewmaketitle

\section{Introduction}
%
%
%
%
\IEEEPARstart{O}{pen-source}
SLAM algorithms and commercial mapping solutions offer highly precise single-session 3D maps, however continual changes in environment over time, render these maps less useful over time. 
Hence, regular map updates via lifelong 3D mapping is crucial for long term robot operation, digital twin simulation, long term planning and to derive insights on spatio-temporal changes in the environment. {Existing works on LiDAR based lifelong SLAM/mapping/localization\cite{ltmapper,graph-pruning,adaptive,lifelong-loc,lifelongLocalization,longterm-mapping} focus on a single or a few modules of the whole pipeline \textit{or} propose tightly integrated solutions for online robot operation/localization, which only work with custom designed algorithms for pose graph pruning/optimization, loop closure detection \& pose estimation and cannot generalize to different sensor setups and target environments.}

\begin{figure}[t]
  \begin{center}

  \includegraphics[width=0.4\textwidth]{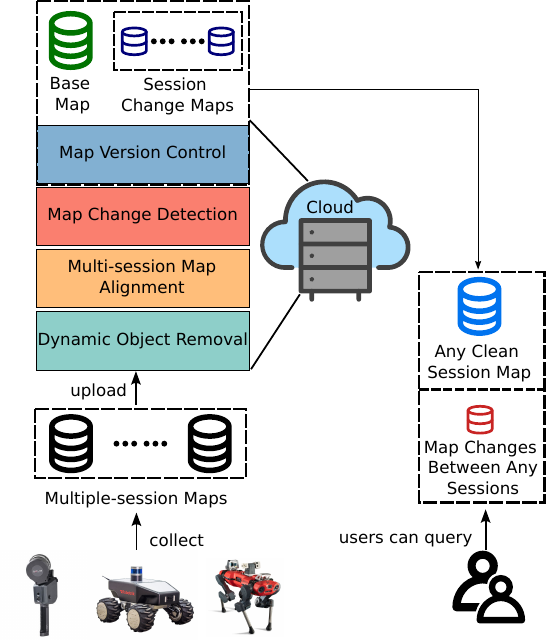} 
  \end{center}
  \vspace{-3mm}
    \caption{Users can upload multi-session 3D maps collected from any source to our proposed lifelong 3D mapping framework. Our system performs dynamic object removal, multi-session map alignment, map change detection and map version control. It allows users to retrieve any clean session map or query changes between any two sessions, all without storing the memory heavy input session maps. 
    }
  \label{fig:Cloud}
   \vspace{-0.7cm}
\end{figure}

Single-session LiDAR 3D maps, ranging from 10GB to 200GB, have three main issues: they contain dynamic objects from the mapping process, each map has a different reference frame due to varying starting poses, and they require substantial storage with huge redundant data. Addressing these challenges, we introduce a sensor-setup agnostic, generic and modular lifelong 3D mapping framework with dynamic point removal, multi-session map alignment, map change detection, and map version control. Our system, depicted in Fig. \ref{fig:Cloud}, processes input session maps with the mentioned artifacts and allows users to retrieve any clean session map and query changes between any two mapped sessions, all without storing memory heavy input session maps, making the system very unique and highly memory efficient.

{We propose a \textit{generic, sensor-setup agnostic} and \textit{modular} 3D lifelong mapping framework that works with both hand-held and robot-mounted LiDAR mapping devices, with these key contributions in increasing order of novelty:}
\newline \textbf{1.} {A generic dynamic point removal method with no sensor-setup or LiDAR motion dependant constraints and a two-stage multi-session map alignment algorithm that finds optimal parameters using grid search over hyper-parameters.}
\newline \textbf{2.} A map change detection that detects positive and negative map changes between two maps by estimating overlapping regions and employing 2D bird-eye-view (BEV) descriptors.
\newline \textbf{3.} A map version control that maintains and updates a base map based on detected positive and negative differences, while enabling the functionalities (i) to reconstruct any previous session maps and (ii) retrieve any inter-session changes, all without storing the original 3D maps.

%

\section{Related Work}
We present related works on each module and the whole pipeline of our proposed lifelong 3D mapping framework.

\textbf{Dynamic Object Removal:} Traditional dynamic object detection methods often rely on handcrafted techniques or geometric and temporal consistency descriptors. Removert\cite{removert} utilizes range image-based point discrepancy calculation to remove potential dynamic objects, while Peopleremover\cite{peopleremover} employs a 3D occupancy voxel grid and ray-casting to identify occupied voxels across multiple scans. ERASOR\cite{erasor} introduces a region-wise pseudo occupancy descriptor and a scan ratio test for dynamic point retrieval, coupled with ground plane fitting for static point differentiation. However, ERASOR and similar handcrafted approaches assume a horizontal LiDAR placement and planar vehicle motion, limiting their applicability to handheld LiDAR mapping. While Octomap\cite{octomap} by default is sensor-setup agnostic, Arora et al.\cite{groundoctomap} enhanced its performance by employing a ground plane detection algorithm that relies on sensor-height \& LiDAR placement on the robot/vehicle, making it unsuitable for hand-held LiDAR mapping devices. Deep learning-based methods like 4DMOS, LM-Net, and MotionSeg3D are tailored for robot-mounted and autonomous driving scenarios, rendering them inflexible to varying sensor setups, type of LiDAR (16-32-64-128 channel) and trained environments.

\textbf{Multi-session Map Alignment:} Kim et al.\cite{relativeMultiSession} proposed a multi-session relative pose graph for faster incremental smoothing and mapping. LT-Mapper\cite{ltmapper} employs ScanContext descriptors to establish inter-session and intra-session loop closures but faces challenges with false loop constraints in complex environments that end up corrupting even the accurate input SLAM 3D maps. For multi-session map alignment, various registration algorithms like ICP, GICP, or NDT can be used, but optimal registration isn't guaranteed. Some researchers have explored 3D keypoint correspondence methods\cite{igsp} for initial alignment using descriptors like FPFH, 3D-HoPD\cite{hopd} or B-SHOT\cite{bshot} before refining registration with ICP, NDT or energy function minimization\cite{igsp}.


\textbf{Map Change Detection:} Stilla et al.\cite{change-review} reviewed major works on urban 3D point cloud change detection, categorizing them into point-based, voxel-based, and segment/object-based methods. Point-based techniques measure spatial differences using metrics like Euclidean distance, projected 2D distance or multiscale model-to-model distance\cite{change-distance-2}. Voxel-based approaches\cite{fast-change} voxelize point clouds and compare voxel states while segment-based methods rely on robust clustering or segmentation algorithms but struggle with generalization. LT-Mapper\cite{ltmapper} detects changes by identifying low dynamic points and detecting strong positive differences, heavily relying on Removert\cite{removert}, which again makes it sensor-setup dependant and unadapatable to hand-held LiDAR mapping devices. Our proposed work stands out as we \textit{detect these changes across multi-session maps} \& \textit{tightly integrate them into a map version control framework}.

\textbf{Map Version Control:} Using Git for versioning two 3D point cloud maps isn't effective due to Git's inability to detect changes in non-textual data. Ogayar-Anguita et al.\cite{anguita} developed a version control system for point clouds but it's limited to single input point cloud and  assumes that if a 3D point's coordinate changes, then its a new change, and every change shall preserve order of points. This design does not work with multi-session maps because of the imminent small alignment errors in multi-session map alignment and multi-session maps can never be ordered.


{Other lifelong SLAM methods include pose graph sparsification\cite{lifelong-loc}, {{seperate lifelong dual-memory}\cite{bioslam}}, {{optimization-based view management}\cite{wildSLAM}}, adaptive local map pruning\cite{adaptive}, filtering dynamic points and adding new regions with constant density\cite{3dmaintenance}, and spatial transfer of semantics and relative constraints\cite{semanticUpdate}. However, these methods are tailored for online robot operation with specific sensor setups, lack modularity and are not sensor-setup agnostic.} LT-Mapper\cite{ltmapper} introduces a modular lifelong 3D mapping framework but relies on Removert\cite{removert} for dynamic object removal and map change detection, limiting its adaptability for hand-held LiDAR devices. ScanContext\cite{scancontext}, used for multi-session map alignment, suffers from false positive matches, compromising map accuracy. Although LT-Mapper provides insights into delta map and map rollback, there's a need for clearer analysis, better modularization, and a sensor-setup agnostic design.

\section{Proposed Lifelong 3D Mapping Framework}
Our method, depicted in Fig. \ref{fig:Cloud}, begins with sensor-setup agnostic dynamic object removal and automatic parameter-free multi-session map alignment. The map change detection module identifies positive and negative differences between aligned maps, which are managed by the map version control module to maintain a base map representing the current state of the environment. Our system offers users the functionality to download any previous clean session maps, or query changes between any two sessions.


\subsection{Illustration of the Complete Workflow }
\label{sec:workflow}

The complete workflow is shown in Fig. \ref{fig:wholePipeline}. The proposed system accepts multiple session maps as input and each 3D map is represented as a set of {poses and point clouds}. 

Let's say $n+1$ $session\ maps(t)$, where $t=0,1,...,n$, are uploaded to the system over time. As shown in Fig. \ref{fig:wholePipeline}(a), the first $session\ map(t_0)$ is processed by dynamic object removal and the resulting clean $session\ map(t_0)$ is used to initialize the first $base\ map(t_0)$. Next, as shown in Fig. \ref{fig:wholePipeline}(b), the existing $base\ map(t)$ and next $session\ map(t+1)$ act as inputs to the system. The $session\ map(t+1)$ is first processed by dynamic object removal to create a clean $session\ map(t+1)$. The multi-session map alignment aligns $base\ map(t)$ and clean $session\ map(t+1)$. These aligned $base\ map(t)$ and $session\ map(t+1)$ are then processed by map change detection, producing $base\ ND(t)$ and $session\ PD(t+1)$, representing the negative and positive differences which are explained in detail in Sec. \ref{Sec:MapChange}. Lastly, the map version control module updates the $base\ map(t)$ to $base\ map(t+1)$ and stores $base\ ND(t)$, $session\ PD(t+1)$ and boundary points of every session map, while freeing the memory of $session\ map(t+1)$. The map version control interfaces with the users and, upon request, it will reconstruct any clean $session\ map(t)$ and return changes between any two clean $session\ maps(t,t+n)$. Let us now dive deep into each of these proposed modules.

\begin{figure}[h]
  \vspace{-2mm}
  \begin{center}
  \includegraphics[width=0.38\textwidth]{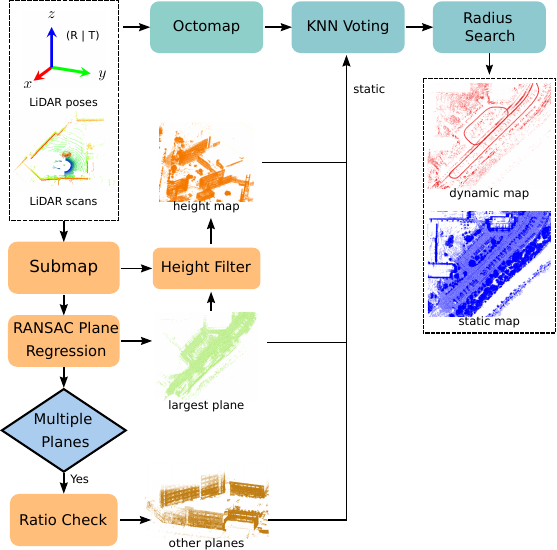} 
  \end{center}
\vspace{-4mm}
  \caption{Our proposed sensor-setup agnostic dynamic point removal pipeline. We create a submap, regress multiple planes and add them back based on a ratio check to fill the residual holes from OctoMap. We perform radial search based post-processing to further improve the quality of static maps. }
  \label{fig:dynamicObjectRemovalPipeline}
  \vspace{-5mm}
\end{figure}

\subsection{Dynamic Object Removal}

We introduce a sensor-setup and motion-agnostic dynamic object removal pipeline as depicted in Fig. \ref{fig:dynamicObjectRemovalPipeline}. {The input LiDAR scans along with their poses are first processed using OctoMap\cite{octomap} to generate a static map, dynamic map and a set of unknown points. 
However, these resulting static and dynamic maps have two main problems. First, there are many residual holes in planar regions due to imprecise probabilities at voxel boundaries, particularly noticeable in ground planes and large building facades. Second, the created static maps are much sparser and contain residual dynamic points.}

{To overcome the first issue, as in Fig. \ref{fig:dynamicObjectRemovalPipeline}, we create a submap by concatenating about 10-50 frames using the poses from LiDAR mapping systems. Then a RANSAC-based plane regression algorithm detects multiple planes in the created submap. By default, in most scenarios, the largest detected plane is directly added to the static map from OctoMap, as this largest detected plane mostly represents the ground plane or a large building facade. We then calculate the ratio of number of points in other detected planes with respect to the number of points in the submap, determining if they are large enough. If this ratio is greater than a threshold, we add the subsequent large planes to the static map. Generally, the largest plane is the ground plane and subsequent planes are mostly large building facades, and adding these planes back to static map, fills the residual holes while keeping the algorithm sensor-setup independent. Optionally, a height filter that classifies all points above a certain height as static can be added, which can greatly reduce the computational requirements. However, this height filter may not directly work in the case of drone-based mapping or hand-held mapping in multi-storey buildings, and thus may require additional consistency checks to use it.}

To address the second issue of sparse static maps and remove remaining dynamic points, we start by categorizing unknown points from OctoMap as static or dynamic using K-NN voting\cite{groundoctomap}. Subsequently, we also apply statistical outlier removal and perform radial filtering to reassign points close to static points in the dynamic map as static, reducing misclassified dynamic points. Additionally, we can classify each point in every scan as a static or dynamic point based on a further K-NN based radius search. Even after this, there will be about 1-3\% of static points misclassified as dynamic points, as all areas cannot be revisited during mapping.


\subsection{Multi-session Map Alignment}


We employ a two-stage point cloud registration, starting with a robust initial alignment via feature descriptor matching and followed by a precise registration using Normal Distributions Transform (NDT)\cite{ndt}. The 3D maps are first down-sampled using the keypoint radius parameter $K_{r}$ to generate keypoints. We then extract the efficient PCA-SHOT\cite{pcashot} 3D feature descriptor for each keypoint, which depends on additional parameters: point cloud down-sampling $PC_{ds}$, normal neighbors $N_{n}$, and descriptor radius ${FD}{r}$. Utilizing the 50-dimensional PCA-SHOT descriptors, an initial alignment estimate is computed using RANSAC, with ransac inlier ratio, set as default to twice of $K{r}$. This alignment is used to trigger the fine registration with NDT, guided by two parameters: resolution $NDT_r$ and step size $NDT_{ss}$. We use the downsampled point cloud based on $PC_{ds}$, as it significantly fastens the NDT registration over using raw point clouds. NDT has shown to be more computationally efficient, robust, and accurate in real-world tests, outperforming alternatives like ICP, GICP, and VoxelizedGICP\cite{vgicp}. Additionally, while alternatives to PCA-SHOT such as PCA-FPFH\cite{pcashot} and B-SHOT\cite{bshot} have been tested, PCA-SHOT provided a greater number of correspondences with fewer keypoints, thus increasing the system's robustness.


The multi-session map alignment pipeline requires six key parameters: $K_{r}$, $PC_{ds}$, $N_{n}$, ${FD}{r}$, $NDT_r$, and $NDT{ss}$. Table. \ref{tab:ParameterSet} recommends a parameter set based on our real-world experiments. We perform iterative grid search in this hyper parameter space. Most parameter subsets fail at the feature matching stage, because of less PCA-SHOT descriptor's correspondences or less inliers in RANSAC based registration. Only those parameter configurations that succeed are advanced to the fine registration using NDT. The optimal alignment is determined by selecting the registration with the lowest Chamfer distance, as detailed in Sec. \ref{EXP-B}.

\begin{table}[htp]
\vspace{-3mm}
\centering
\caption{Parameter lists used in all experiments}
\resizebox{.35\textwidth}{!}{%
\begin{tabular}{l|c}
\hline
Parameter                                     & Value Range             \\ \hline
keypoint radius(in m) - $K_{r}$ & {[}0.5,1,2,5,10{]}      \\
point cloud down-sampling(in m) - $PC_{ds}$    & {[}0.1,0.2,0.3,1{]}     \\
normal neighbours - $N_{n}$                                 & {[}200,500{]}           \\
feature descriptor radius(in m) - $FD_{r}$                                   & {[}5,10,20,50{]}        \\
NDT resolution(in m) - $NDT_{r}$                                & {[}0.5,1,2,5{]}         \\
NDT step size - $NDT_{ss}$                                 & {[}5,10{]}              \\ \hline
\end{tabular}%
}
\label{tab:ParameterSet}
\vspace{-7mm}
\end{table}

\subsection{Map Change Detection}\label{Sec:MapChange}

\begin{figure}[htbp]
  \vspace{-2mm}
  \begin{center}
  \includegraphics[width=0.35\textwidth]{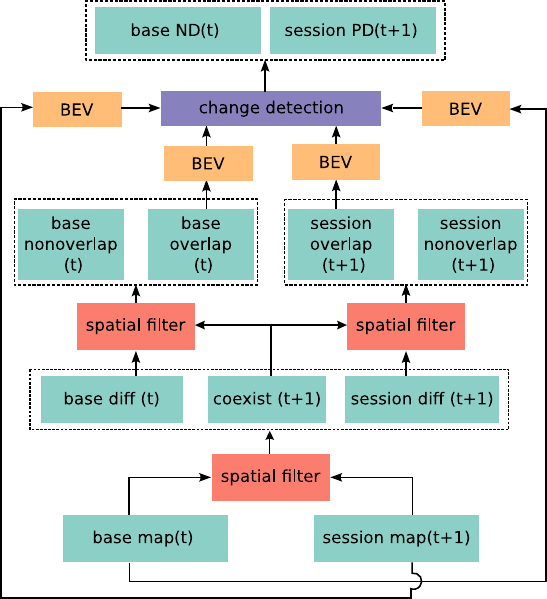} 
  \end{center}
\vspace{-3mm}
  \caption{Map Change Detection - Given $base\ map($t$)$ and $session\ map(t+1)$, the map change detection identifies negative differences $base\ ND(t)$ and positive differences $session\ PD(t+1)$. Negative difference is the objects that were present in base map but disappeared in session map while the positive difference is the map data that newly appeared in session map.}
  \label{fig:Map-change-detection-pipeline}
  \vspace{-3mm}
\end{figure}

The first clean/static session map is used as the initial base map. Map change detection identifies the changes between the base map and incoming session map. As illustrated in Fig. \ref{fig:Map-change-detection-pipeline}, it takes as input, $base\ map(t)$ and $session\ map(t+1)$ and outputs the negative changes in $base\ map(t)$ as $base\ ND(t)$ and the positive changes in $session\ map(t+1)$ as $session\ PD(t+1)$. The $base\ ND(t)$ denotes the 3D map data that was present in $base\ map(t)$ but disappeared in $session\ map(t+1)$, while $session\ PD(t+1)$ indicates the new 3D data that appeared in $session\ map(t+1)$ and was absent in $base\ map(t)$.

Initially, as depicted in Fig. \ref{fig:Map-change-detection-pipeline}, $base\ map(t)$ and $session\ map(t+1)$ are processed by a k-NN-based spatial filter to create $base\ diff(t)$, $coexist(t+1)$, and $session\ diff(t+1)$. A radius search is performed for each $base\ map(t)$ point in $session\ map(t+1)$, classifying points with radial neighbors as $coexist(t+1)$ and those without as $base\ diff(t)$. We repeat the same for points in $session\ map(t+1)$, searching in $base\ map(t)$ to find $coexist(t+1)$ and $session\ diff(t+1)$. This yields two similar sets of $coexist(t+1)$; either can be selected.

Next, another spatial search is performed for each $base\ diff(t)$ point in $coexist(t+1)$, classifying points with radial neighbors as $base\ overlap(t)$ and those without as $base\ nonoverlap(t)$. We repeat this process for $session\ diff(t+1)$ and $coexist(t+1)$ generating $session\ overlap(t+1)$ and $session\ nonoverlap(t+1)$. 


{BEV descriptor compresses spatial information of the 3D point cloud into 2D matrices, thus accelerating the comparison for map change detection. To realize this, we first project the 3D point clouds into 2D canonical planar grid along detected normal direction using PCL library's plane regression. This 2D planar grid is converted into an image with each pixel intensity representing the maximum height of 3D points that fall in that specific grid, creating a 2D Bird Eye View (BEV) descriptors. 
The negative difference is estimated by comparing the two 2D BEV descriptors of $session\ map(t+1)$ and $base\ overlap(t)$, resulting in changes at pixel level granularity/spatial resolution. The final detected change comprises of 3D points that fall in the pixels with significant difference in their intensity values. The same procedure applies to $session\ map(t+1)$ and $base\ overlap(t)$ to identify the positive difference. For precise change detection, a resolution of 0.05-0.15m is recommended, while 0.5m-2m is suggested for high-efficiency on large 3D maps.}

Our notation categorically associates negative differences (ND) with the current base map as $base\ ND(t)$ and positive differences(PD) with the incoming session map as $session\ PD(t+1)$, aligning them with their time indexes, $t$ and $t+1$, for intuitive understanding and consistent version control. In indoor car-parking scenarios, multi-layered 2D BEV descriptors are used to detect changes across layers.  

\subsection{Map Version Control}

The map version control system has two main functions: it can reconstruct any former clean session map and provide changes between any two chosen sessions without storing the input session maps. It takes as input, the positive and negative differences from the map change detection and updates the base map to represent the current state of the environment. The system stores these positive and negative differences to facilitate forward progression of the map version control, i.e., to update the base map, as outlined in the equations below. Additionally, to allow for the reconstruction of previous session maps, it calculates and saves the boundaries of every new session map using the convex hull filter from the Point Cloud Library (PCL).

For the forward progression step, the $base\ map(t)$ is updated to $base\ map(t+1)$ as:
\vspace{-0.2cm}
\begin{equation*}%
\begin{split}%
base\ map(t+1) =&\ coexist(t+1) + base\ overlap(t)\ \ \\
+\ base\ nonoverlap(t)& + session\ nonoverlap(t+1)\ +\\ session\ PD(t+1)& - base\ ND(t)
\end{split}
\end{equation*}
where $+$ represents point cloud concatenation and $-$ represents point cloud removal, which can be realized using k-NN based radial search with very small radius. Please note that once the $base\ map(t)$ is updated to $base\ map(t+1)$, except the positive and negative differences, all the used information in the above equation such as $session\ map(t+1)$, $base\ overlap(t)$, $base\ nonoverlap(t)$ and $session\ nonoverlap(t+1)$ are not needed anymore.

Next, to reconstruct previous session maps, i.e., in order to perform backward progression of the map version control, we only require the current base map, saved positive and negative differences, and corresponding boundaries of each session map. This can be realized as:

\begin{footnotesize}
\begin{equation}
\label{eqn1}
   \mathcal{M}^{'}_s(k)=HullFilter^{k}\{\mathcal{M}_b(t)+\sum_{i=t}^{k} (base\ ND(i-1)- session\  PD(i)) \}
\end{equation}
\end{footnotesize}

where $\mathcal{M}^{'}_s(k)$ is the reconstructed session map at time $k$, $\mathcal{M}_b(t)$ is the current base map at time $t$, \mbox{$base\ ND(i-1)$} and $session\ PD(i)$ are the stored negative and positive differences at specific time stamps and lastly $HullFilter^{k}$ is the convex hull filter that takes in the stored session map's boundary points at time $k$ and operate/crop the base map to reproduce the desired session map. Lets look at an example.

\textbf{Example:} A user uploads four session maps collected at different times ($t_0, t_1, t_2, t_3$) and after processing, the map version control module contains the latest base map, $base\ map(t_3)$, negative and positive changes \{$base\ NDs(t_0,t_1,t_2)$\} and \{$session\ PDs(t_1,t_2,t_3)$\}, and boundaries of each session map $session\ BPts(t_0,t_1,t_2,t_3)$. 

If a user requests for a previous $session\ map(t_1)$, then it can be reconstructed using Eqn.(\ref{eqn1}) as:
\vspace{-2mm}

\begin{footnotesize}
\begin{equation*}
\begin{split}
session\ map(t_1) = HullFilter^{t_1}\{base\ map(t_3) + \\base\ ND(t_2) - session\ PD(t_3) + base\ ND(t_1) - session\ PD(t_2)\}
\end{split}
\end{equation*}
\end{footnotesize}

\textbf{Explanation:} In the above equation, when we add $base\ ND(t_2)$ and remove $session\ PD(t_3)$ to the latest $base\ map(t_3)$, then $base\ map(t_3)$ becomes equivalent to $base\ map(t_2)$. And then when we add $base\ ND(t_1)$ and remove $session\ PD(t_2)$, the base map changes to $base\ map(t_1)$. This $base\ map(t_1)$ represents the state of the environment at time $t_1$, and we crop the region that belongs to session map by applying the $HullFilter^{t_1}$ filter with $session\ BPts(t_1)$.

If the user requests for changes between $session\ map(t_1$) and $session\ map(t_2$), then we reconstruct both these session maps using the above step and pass the reconstructed maps to the map change detection module to retrieve map changes. Additionally, we can also save the boundary points of every base map to enable further additional features that require precise reconstruction of previous base maps.

\begin{figure}[t]
  \begin{center}
  \includegraphics[width=0.46\textwidth]{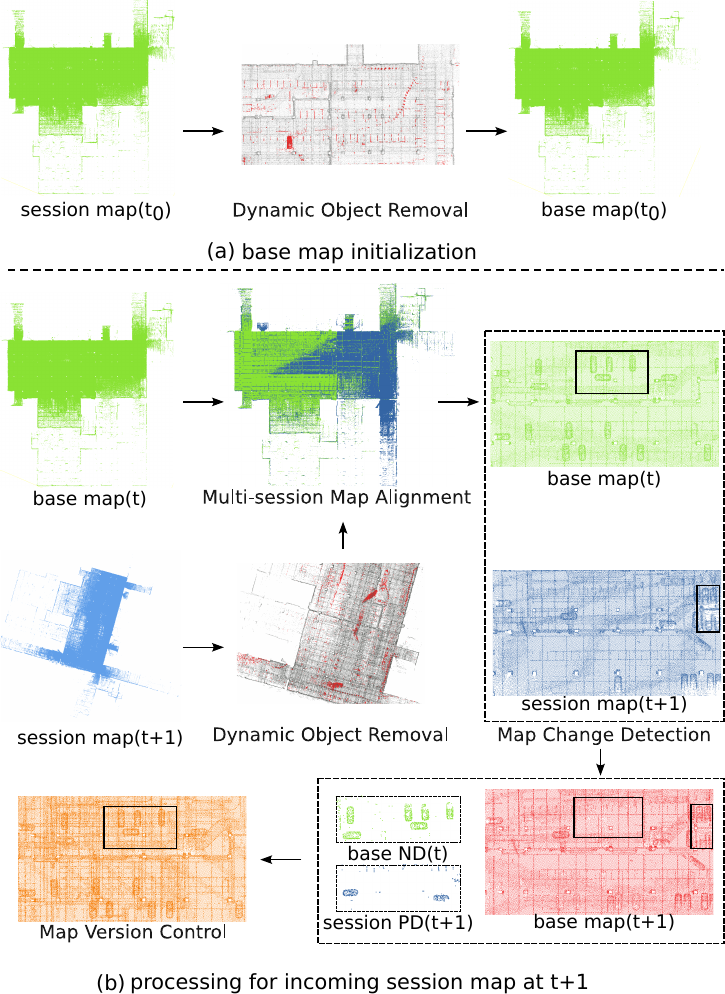} 
  \end{center}
  \vspace{-3mm}
  \caption{\textbf{Illustration of Complete Workflow}. (\textbf{a}) Initialization of base map. {\color{gray}Grey} points are static and {\color{red}red} points are dynamic. \textbf{(b)} $session\ map(t+1)$ first goes through dynamic object removal and is then aligned to $base\ map(t)$. The map change detection detects negative changes $base\ ND(t)$ and positive changes $session\ PD(t+1)$ while the map version control module updates the $base\ map(t)$ to $base\ map(t+1)$. The rectangular boxes highlight regions with changes and how the changes in $session\ map(t+1)$ get reflected in $base\ map(t+1)$. Specifically, the partly occluded region between parked cars, as highlighted in rightmost box in session map ($t+1$) is maintained as before, in updated $base\ map(t+1)$. }
  \label{fig:wholePipeline}
  \vspace{-5mm}
\end{figure}

\begin{figure}[tp]
  \begin{center}
  \includegraphics[width=0.42\textwidth]{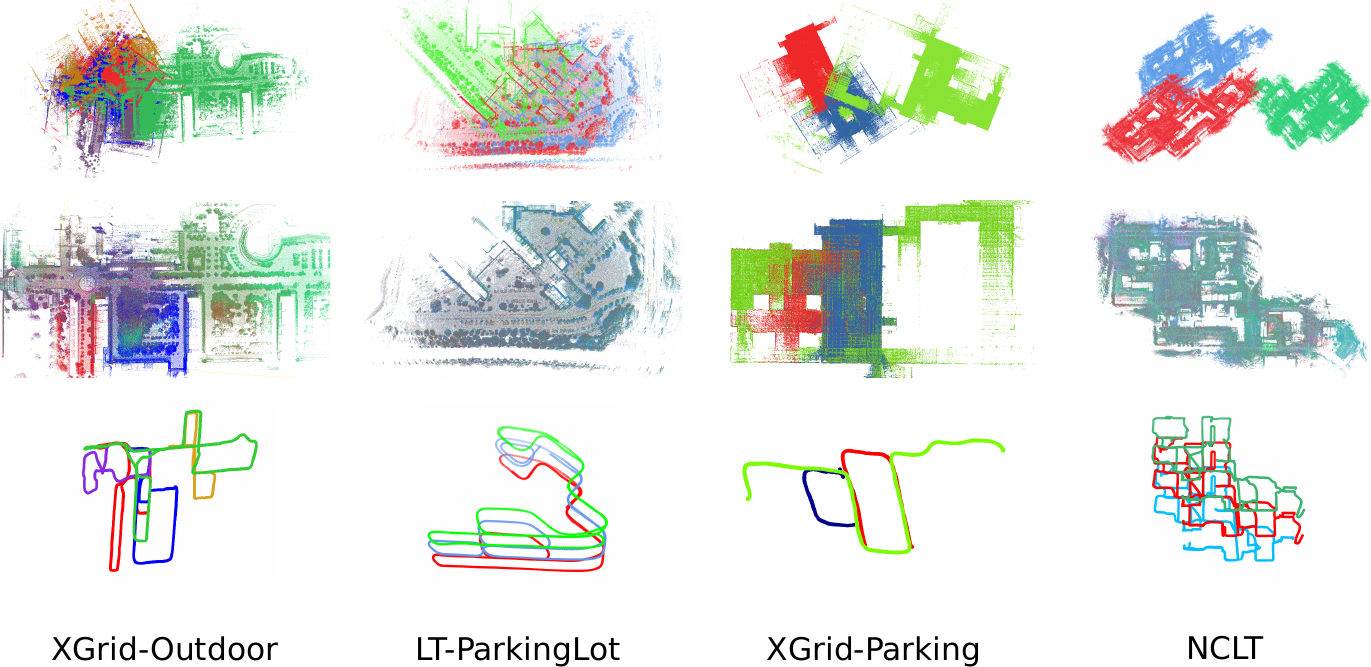} 
  \end{center}
  \vspace{-4mm}
  \caption{Multi-Session map alignment results on multiple datasets with different local coordinate frames. Height difference is added to visualize \& distinguish different trajectories.}
  \label{fig:multi-session-map-alignment}
  \vspace{-4mm}
\end{figure}

\section{Experiments}
We performed extensive evaluation of each module and the full pipeline on diverse open-source datasets including SemanticKITTI\cite{semantickitti}, NCLT\cite{nclt}, LT-Parking Lot\cite{ltmapper}, {MulRan \cite{Mulran}} and commercial hand-held LiDAR mapping device, XGrids, over long periods of time covering many kilometers.

\subsection{Dynamic Object Removal}

\begin{table}[htbp]
\centering
\caption{Dynamic object removal results on SemanticKITTI}
\resizebox{0.35\textwidth}{!}{%
\begin{tabular}{ccccc}
\hline
Seq                 & Method         & PR              & RR              & F1              \\ \hline
\multirow{4}{*}{00} & ERASOR         & 0.9172          & 0.9700          & 0.9429          \\
                    & Removert       & 0.9328          & 0.7663          & 0.8414          \\
                    & Ground-Octomap & 0.7765          & 0.9526          & 0.8556          \\
                    & Ours           & \textbf{0.9471} & \textbf{0.9712} & \textbf{0.9590} \\ \hline
\multirow{4}{*}{01} & ERASOR         & 0.9193          & 0.9463          & 0.9326          \\
                    & Removert       & \textbf{0.9579} & 0.6688          & 0.7877          \\
                    & Ground-Octomap & 0.8475          & 0.7337          & 0.7865          \\
                    & Ours           & 0.9425          & \textbf{0.9528} & \textbf{0.9477} \\ \hline
\multirow{4}{*}{02} & ERASOR         & 0.8108          & \textbf{0.9911} & 0.8919          \\
                    & Removert       & 0.8531          & 0.8222          & 0.8374          \\
                    & Ground-Octomap & \textbf{0.9479} & 0.6277          & 0.7553          \\
                    & Ours           & 0.9421          & 0.9035          & \textbf{0.9224} \\ \hline
\multirow{4}{*}{05} & ERASOR         & 0.8698          & \textbf{0.9788} & 0.9211          \\
                    & Removert       & 0.9223          & 0.6757          & 0.7800          \\
                    & Ground-Octomap & 0.6956          & 0.9390          & 0.7992          \\
                    & Ours           & \textbf{0.9425} & 0.9712          & \textbf{0.9566} \\ \hline
\multirow{4}{*}{07} & ERASOR         & 0.9200          & \textbf{0.9833} & 0.9506          \\
                    & Removert       & 0.8482          & 0.5758          & 0.6860          \\
                    & Ground-Octomap & 0.5396          & 0.9081          & 0.6769          \\
                    & Ours           & \textbf{0.9768} & 0.9410          & \textbf{0.9586} \\ \hline
\multirow{4}{*}{Mean} & ERASOR         & \multicolumn{1}{r}{0.8874}          & \multicolumn{1}{r}{\textbf{0.9739}} & \multicolumn{1}{r}{0.9278}          \\
                      & Removert       & \multicolumn{1}{r}{0.9029}          & \multicolumn{1}{r}{0.7017}          & \multicolumn{1}{r}{0.7865}          \\
                      & Ground-Octomap & \multicolumn{1}{r}{0.7614}          & \multicolumn{1}{r}{0.8322}          & \multicolumn{1}{r}{0.7747}          \\
                      & Ours           & \multicolumn{1}{r}{\textbf{0.9502}} & \multicolumn{1}{r}{0.9479}          & \multicolumn{1}{r}{\textbf{0.9488}} \\ \hline
\end{tabular}%
}
\vspace{1ex}

{\centering PR denotes preservation rate, RR denotes rejection rate. \par}
\label{tab:dynamicObjectRemoval4Compare}
  \vspace{-7mm}
\end{table}

We perform quantitative analysis on SemanticKITTI\cite{semantickitti} dataset, which offers point-wise ground truth labels. Points from moving categories are classified as dynamic, while the rest are considered static. Following ERASOR\cite{erasor}, we use preservation rate ($PR$), rejection rate ($RR$), and the $F1$ score for evaluation. $PR$ measures the model's ability to keep static points, $RR$ gauges the proportion of correctly identified dynamic points, and the $F1$ score is the harmonic mean of $PR$ and $RR$, indicating overall performance. The sequence and scan range chosen for evaluation on SemanticKITTI match ERASOR's\cite{erasor} experimental setup. 

Table \ref{tab:dynamicObjectRemoval4Compare} compares the performance of our method with state-of-the-art methods such as ERASOR\cite{erasor}, Ground-Octomap\cite{groundoctomap}, and Removert\cite{removert}. Our method achieves the highest mean preservation rate of 0.95, which is at least 5-7\% better than other methods, ensuring that the static maps retain more static points — the core objective of any dynamic object removal algorithm. While our rejection rate is roughly 3\% lower than ERASOR's, it's important to note that ERASOR used a lower frame rate, which boosted their metrics but resulted in sparser maps. As for the overall classification accuracy based on F1 score, our proposed method offers 2-3\% improvement over state-of-the-art.



\begin{figure*}[ht]
  \begin{center}
  \includegraphics[width=0.85\textwidth]{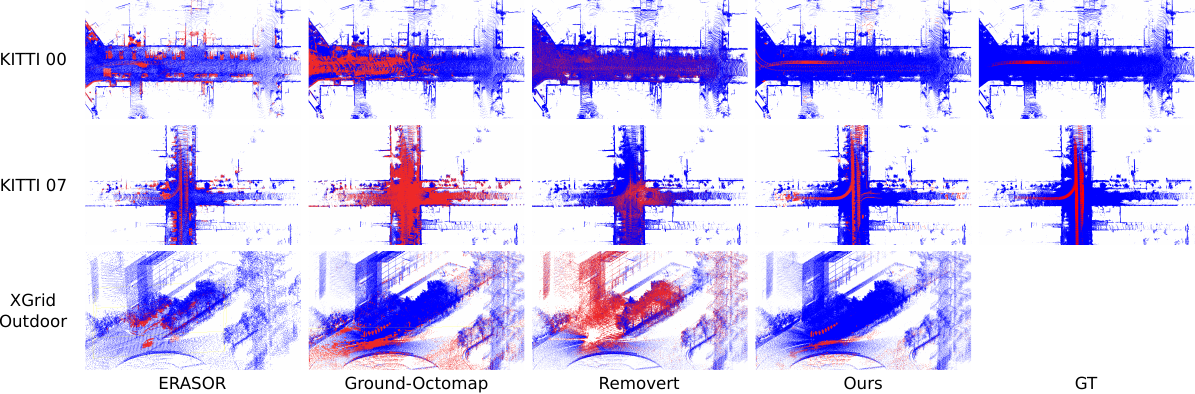} 
  \end{center}
\vspace{-4mm}
  \caption{Dynamic object removal on SemanticKITTI and XGrid dataset. {\color{blue}Blue} points are static points and {\color{red}red} points are dynamic points. A close zoom-in clearly shows the our proposed method performs better than ERASOR and clearly removes human trajectories and dynamic objects on X-Grid-Outdoor dataset. ERASOR cannot process every frame because of high memory and compute requirements, resulting in much sparser maps on every dataset.}
  \label{fig:dynamicObjectRemovalComparison}
  \vspace{-3mm}
\end{figure*}

We show qualitative results on two KITTI sequences and one outdoor sequence from XGrids LiDAR mapping device in Figure \ref{fig:dynamicObjectRemovalComparison}. Visually, on all three sequences, it is evident that Ground-Octomap and Removert detect significantly high erroneous dynamic points, shown in red color. The drawbacks of ERASOR method is that they assume that dynamic objects are in contact with the ground plane and to detect the ground plane, they use the lowest height points as seeds points for plane estimation, which does not always hold true with hand-held LiDAR mapping devices, as their motion is unconstrained. Hence, zooming into the results of ERASOR on XGrid-Outdoor dataset in Fig. \ref{fig:dynamicObjectRemovalComparison} reveals that it detects many points on static trees as dynamic. On the same XGrid-Outdoor dataset in Fig. \ref{fig:dynamicObjectRemovalComparison}, Removert gives completely erroneous results, while Ground-Octomap detects large ground regions as dynamic points. A close look reveals that our proposed method precisely detects a walking human and a dynamic object moving on the road, while offering more accurate and dense static map on all sequences. 

These mainstream algorithms, range-image based Removert\cite{removert}, the ground estimation based Ground-Octomap\cite{groundoctomap} and ERASOR\cite{erasor} were mainly developed for autonomous driving scenarios, rendering them less accurate for other use cases. {Our proposed methods requires approximately 1.5 hours to process the whole KITTI 00 sequence with ~4500 frames, which is 1.5x faster than Ground-Octomap on our PC with AMD Ryzen 9 3900x CPU. ERASOR's computational and memory requirements were so higher that it could only work with 4x lower frames and could not process every frame, resulting in sparse maps.}





\begin{figure}[!]
  \vspace{-1mm}
  \begin{center}
  \includegraphics[width=0.48\textwidth]{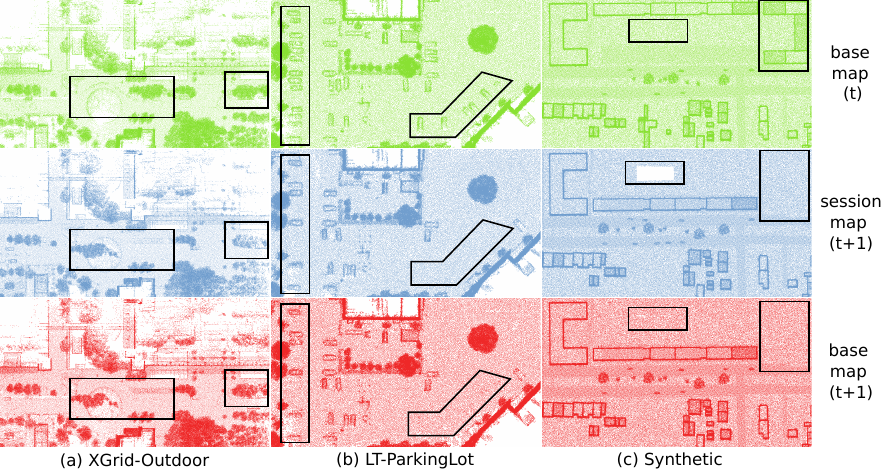} 
  \end{center}
    \vspace{-5mm}
  \caption{Visualization of the map change detection on XGrid-Outdoor, LT-ParkingLot and Synthetic datasets. In (a), the left rectangle shows that our module removed the disappeared trees while adding the newly grown trees in $base\ map(t+1)$. The right rectangle shows that the useful information occluded in session map ($t+1$) is maintained as is, in the updated base map ($t+1$). In (b), the two highlighted boxes show how the changes in car parking in $session\ map(t+1)$ get updated onto the $base\ map(t+1)$. In (c), in left rectangle, we simulate a scenario where the highlighted region is occluded and not sensed in session map ($t+1$). Our algorithm preserves and maintains the original map from base map ($t$). In right rectangle, an existing building in $base\ map(t)$ disappears in $session\ map(t)$, which successfully gets updated in latest $base\ map(t+1)$. }
  \label{fig:map-change-detection-result}
  \vspace{-4mm}
\end{figure}

\subsection{Multi-session Map Alignment}
\label{EXP-B}

The Chamfer distance is one of the most widely used metric for evaluating point cloud registration, and it is the mean of squared distances between nearest neighbor correspondences of two point clouds. 
\begin{footnotesize}
\begin{multline*}
   CD(X',Y')=\frac{1}{|X|}\sum_{x \in X} \min_{y\in Y'}\|x-y\|_2^2+\frac{1}{|Y|}\sum_{y \in Y} \min_{x\in X'}\|x-y\|_2^2 
\end{multline*}
\end{footnotesize}

where $X'$ and $Y'$ are the raw point cloud sets, $X=\{x\in X'| \exists y\in Y', \: \text{s.t.} \: \text{min}\|x-y\|_2<\tau \}$ and $Y=\{y\in Y'|\exists x\in X', \: \text{s.t.} \: \text{min} \|x-y\|_2<\tau \}$. A distance threshold $\tau$, set to 0.5 was used to filter outliers that do not have nearest neighbours.


{Our two-stage multi-session map alignment was benchmarked against the LT-Mapper's multisession map alignment, ICP and NDT approach on four datasets: LT-ParkingLot, XGrid-Outdoor, MulRan and XGrid-Parking datasets}. {XGrid-Outdoor and XGrid-Parking's poses were retrieved from XGrids's proprietary software, while SC-LIO-SAM and SC-A-LOAM were used to get the poses for the LT-ParkingLot and MulRan dataset. In our testing, we aligned 6 maps from the LT-ParkingLot dataset, 5 maps from XGrid-Parking,} {3 maps from MulRan DCC sequence and 6 maps from XGrid-Outdoor.} The average Chamfer distance post-alignment is shown in {Table \ref{tab:multi-align}}. Our method outperformed others on all datasets. Notably, LT-SLAM struggled with the XGrid-Parking dataset due to the inability of ScanContext to recognize loops within the repetitive structure of the indoor parking environment, failing to establish loop constraints. {The direct alignment via NDT and ICP renders larger-error result, which suffers from the low-density distinctive information in large-scale maps.}

Qualitative results are shown in Fig. \ref{fig:multi-session-map-alignment}, displaying a selection of maps for clarity. Our multi-session map alignment's efficacy is also demonstrated on the NCLT dataset, collected over a year, showcasing its robustness and adaptability across diverse environments. {To align the maps from XGrid Outdoor with about 10M points, our method takes roughly 3 mins to do all the grid search. In comparison, LT-SLAM has to traverse all the keyframe descriptors for loop detection and graph optimization, which takes over 10 mins.}

\begin{table}[]
\caption{{Chamfer distance after multi-session map alignment}}
\centering
\resizebox{0.4\textwidth}{!}{%
\begin{tabular}{ccccc}
\hline
Dataset       & Ours            & LT-SLAM &{ICP} &{NDT} \\ \hline
XGrid-Outdoor & \textbf{0.0690} & 0.0862 & {0.1389} & {0.1342} \\
XGrid-Parking & \textbf{0.0812} & fail   & {fail} & {0.1235}\\
LT-ParkingLot-SC-LIO-SAM & \textbf{0.0329} & 0.0592 &{0.1211} &{0.1378} \\ 
LT-ParkingLot-SC-A-LOAM & \textbf{0.0173} & 0.0526  &{0.1228} &{0.1282} \\ 
{MulRan-SC-LIO-LOAM} & {\textbf{0.1004}} & {0.2059}  &{fail} &{0.1231} \\
\hline
\end{tabular}%
}
\vspace{1ex}
\label{tab:multi-align}
  \vspace{-5mm}
\end{table}

\subsection{Map Change Detection \& Map Version Control}


The map change detection algorithm is designed to remove objects that have disappeared, incorporate newly appeared ones, handle effectively occluded regions, and preserve unchanged areas. To evaluate these capabilities, we used the XGrid-Parking dataset with real car-parking changes, the XGrid-Outdoor dataset with manual and simulated changes, LT-ParkingLot with real-world changes, and a fully Synthetic dataset with occlusions, showcased in Fig. \ref{fig:map-change-detection-result}.

\textbf{Qualitative Evaluation:} The step-by-step process of the proposed lifelong 3D mapping pipeline is illustrated in Fig. \ref{fig:wholePipeline}, using the XGrid-Parking dataset, and discussed in Sec. \ref{sec:workflow}. In Fig. \ref{fig:wholePipeline}(b), the updated $base\ map(t+1)$ excludes cars that are not present in $session\ map(t+1)$ but includes the new ones, as indicated by the annotated rectangular boxes.

Fig. \ref{fig:map-change-detection-result} presents qualitative results from the XGrid-Outdoor, LT-ParkingLot, and Synthetic datasets, demonstrating the algorithms' effectiveness. In Fig. \ref{fig:map-change-detection-result}(a) and (b), the $base\ map(t+1)$ accurately reflects changes from $session\ map(t+1)$, removing disappeared and adding newly appeared objects as marked by the rectangular boxes. In the case of Synthetic dataset in Fig. \ref{fig:map-change-detection-result}(c), the algorithm not only updates changes but also retains pre-existing map data from $base\ map(t)$ in regions where $session\ map(t+1)$ lacks information due to occlusion.

\begin{table}[]
\centering
\caption{{Precision and recall of map change detection}}
\resizebox{0.35\textwidth}{!}{%
\begin{tabular}{cccccc}
\hline
Dataset                        & Method & PD pr.         & PD re.         & ND pr.         & ND re.         \\ \hline
\multirow{3}{*}{XGrid-Outdoor} & KNN    & 0.747          & 0.933          & 0.738          & \textbf{0.937} \\
                               & PCL-OC & 0.559          & 0.942          & 0.590          & 0.921          \\
                               & Ours   & \textbf{0.968} & \textbf{0.953} & \textbf{0.909} & 0.798          \\ \hline
\multirow{3}{*}{LT-ParkingLot} & KNN    & 0.735          & 0.910          & 0.947          & 0.852          \\
                               & PCL-OC & 0.624          & 0.827          & 0.882          & 0.757          \\
                               & Ours   & \textbf{0.982} & \textbf{0.917} & \textbf{0.952} & \textbf{0.895} \\ \hline
\multirow{3}{*}{XGrid-Parking} & KNN    & 0.524          & 0.667          & 0.555          & \textbf{0.872}          \\
                               & PCL-OC & 0.739          & 0.684          & 0.731          & 0.723          \\
                               & Ours   & \textbf{0.819} & \textbf{0.690} & \textbf{0.945} & 0.848 \\ \hline
\multirow{3}{*}{MulRan DCC}    & KNN    & \textbf{0.836} & 0.810          & 0.717          & 0.756          \\
                               & PCL-OC & 0.695          & 0.686          & 0.564          & 0.798          \\
                               & Ours   & 0.769          & \textbf{0.849} & \textbf{0.875} & \textbf{0.860} \\ \hline
\multirow{3}{*}{Mean}          & KNN    & 0.711          & 0.830          & 0.739          & \textbf{0.854} \\
                               & PCL-OC & 0.654          & 0.785          & 0.692          & 0.800          \\
                               & Ours   & \textbf{0.885} & \textbf{0.852} & \textbf{0.920} & 0.850          \\ \hline
\end{tabular}}
\vspace{1ex}

{\centering PD denotes positive difference($session\ PD(t+1)$), ND denotes negative difference($base\ ND(t)$), pr. represents precision and re. represents recall. \par}
\label{tab:mapChangeDetection}
\vspace{-6mm}
\end{table}
\textbf{Quantitative Evaluation:} We manually introduce changes in XGrid-Outdoor, XGrid-Parking, {MulRan} and LT-ParkingLot datasets by randomly relocating objects such as buildings, cars, and trees, so that we know the ground truth differences/changes. These ground truth changes can then be used to calculate precision and recall for positive ($session PD(t+1)$) and negative differences($base ND(t)$) as defined below:

\begin{footnotesize}
\begin{equation*}
    precision=\frac{\#(\text{true difference points})}{\#(\text{total detected different points})}
\end{equation*}
\begin{equation*}
    recall=\frac{\#(\text{true difference points})}{\#(\text{total ground truth points})}
\end{equation*}
\end{footnotesize}
If the detected difference point contains a ground truth change in its small radial neighborhood, then it is considered as true change. {Table. \ref{tab:mapChangeDetection} shows the precision and recall results for PD($session PD(t+1)$) and ND($base\ ND(t)$) on four datasets, in comparison with K-D tree change detector (KNN) and Octree change detector from Point Cloud Library (PCL-OC), quantifying the performance of map change detection. KNN detects the changes by checking if there exists a 3D point in target map, for every point in source map, within a defined threshold. For PCL-OC, we directly used the implementation from PCL to identify the changed voxels. Our method's precision of PD and ND is significantly higher compared to these two methods. High precision demonstrates that our method can detect more correct PD and ND points while comparable recall means that most of the ground truth changes are correctly identified as PD and ND.} {The map change detection module takes about 2 mins, similar to PCL-OC and KNN, to execute on LT-ParkingLot dataset with simulated changes containing roughly 10K simulated change points in a 3D map with 0.8 million points.}

\vspace{-0.1cm}
\textbf{Memory Efficiency:} The map version control module efficiently manages memory as the number of input multi-session maps grow. We evaluate on 3 multi-session maps from XGrid-Outdoor and XGrid-Parking datasets, 6 maps from LT-ParkingLot and 27 maps from NCLT dataset, collected over 1.5 years. In Table {\ref{tab:memory-map-change-detection}}, we report the memory required to store all input downsampled maps versus memory utilized by the map version control module that essentially saves the base map, negative \& positive differences and boundaries. Our proposed system significantly reduces memory usage compared to storing all input maps, as memory efficiency increases from 40\% to 78\% with 3 to 6 mapping sessions and reaches up to 94\% over a year-long real-world mapping dataset. This efficiency is crucial for long-term robot operation and semantic insights in lifelong maps.

\begin{table}[]
\centering
\caption{{Memory Efficiency of Map Verion Control }}
\resizebox{0.4\textwidth}{!}{%
\begin{tabular}{cccc}
\hline
Dataset(No. of maps)       & All maps (MB) & Ours(MB) & Efficiency ratio \\ \hline
XGrid-Outdoor (3) & 45.4          & 27.2     & 40.1\%             \\
XGrid-Parking (3) & 64.3          & 31.9    & 50.4\%             \\
LT-ParkingLot (6) & 594.1          & 129.9     & 78.1\%            \\ 
NCLT (27) & {7440.9}        &  430.7     & {94.2}\%       \\ \hline
\end{tabular}%
}
\vspace{1ex}

{\centering All point clouds in NCLT dataset were downsampled to 0.5m while other datasets are downsampled to 0.2m. \par}
\label{tab:memory-map-change-detection}
  \vspace{-7mm}
\end{table}

\section{Conclusion}

\vspace{-0.1cm}
We introduced a generic, modular and sensor-setup agnostic lifelong 3D mapping framework enabling retrieval of clean session maps and querying of map changes without storing memory heavy single-session 3D maps. Our framework includes dynamic point removal, multi-session map alignment, map change detection, and version control modules, suitable for both hand-held and robot-mounted LiDAR mapping devices. Our dynamic object removal module creates clean static maps, which are aligned via automatic multi-session map alignment. Map change detection identifies positive and negative changes, that work in tandem with map version control to enable Git-style efficient 3D map management. {Future plans involve applying open-world object detection  and multi-modal foundational modules such as vision language models and LLMs to derive data insights and semantic understanding from lifelong 3D maps.}
\vspace{-0.2cm}


%

\ifCLASSOPTIONcaptionsoff
  \newpage
\fi

\bibliographystyle{unsrt} 
\bibliography{refs}

\end{document}